\pdfoutput=1

\documentclass[11pt]{article}

\usepackage{ACL2023}

\usepackage{times}
\usepackage{latexsym}

\usepackage[T1]{fontenc}

\usepackage[utf8]{inputenc}

\usepackage{microtype}

\usepackage{inconsolata}
\usepackage{amsmath}
\usepackage{hyperref}
\usepackage[linesnumbered,ruled,vlined]{algorithm2e}

\usepackage{xcolor}
\newcommand\MODEL{\textbf{Zebra}}
\newcommand\llama{\textsc{Llama}}

\usepackage{graphicx}
\usepackage{subcaption}
\usepackage{mwe}
\usepackage{multirow}
\usepackage{siunitx}
\usepackage[most]{tcolorbox}
\usepackage{pifont}
\usepackage{color,soul}

\newtcolorbox{example}[2][]{
  enhanced,
  attach boxed title to top left={yshift=-2ex,xshift=4ex},
  colframe=black,
  colback=white,
  fonttitle=\bfseries, 
  colbacktitle=white,
  coltitle=black,
  boxed title style={
    boxrule=0pt,
    colframe=white,
    },
  title=#2,
  #1}

\title{Zebra: Extending Context Window with Layerwise Grouped \\ Local-Global Attention}

\author{
    Kaiqiang Song$^{*}$, Xiaoyang Wang$^{*}$, Sangwoo Cho$^{*}$, Xiaoman Pan, Dong Yu \\
    Tencent AI Lab, Seattle\\
    \{riversong, shawnxywang, swcho, xiaomanpan, dyu\}@global.tencent.com
}

\begin{document}
\maketitle
\renewcommand{\thefootnote}{\fnsymbol{footnote}}
\footnotetext[1]{Equal Contribution}
\renewcommand{\thefootnote}{\arabic{footnote}}

\begin{abstract}
This paper introduces a novel approach to enhance the capabilities of Large Language Models (LLMs) in processing and understanding extensive text sequences, a critical aspect in applications requiring deep comprehension and synthesis of large volumes of information.
Recognizing the inherent challenges in extending the context window for LLMs, primarily built on Transformer architecture, we propose a new model architecture, referred to as \MODEL. 
This architecture efficiently manages the quadratic time and memory complexity issues associated with full attention in the Transformer by employing grouped local-global attention layers.
Our model, akin to a zebra's alternating stripes, balances local and global attention layers, significantly reducing computational requirements and memory consumption.
Comprehensive experiments, including pretraining from scratch, continuation of long context adaptation training, and long instruction tuning, are conducted to evaluate the \MODEL's performance.
The results show that \MODEL~achieves comparable or superior performance on both short and long sequence benchmarks, while also enhancing training and inference efficiency.

\end{abstract}
\section{Introduction}
\label{sec: intro}
To effectively leverage the power of Long Context in Large Language Models (LLMs), it is essential to develop and refine techniques that enable these models to process and interpret extensive text sequences accurately.
This capability is particularly significant in applications that demand deep understanding and synthesis of large volumes of information, such as summarization~\cite{huang-etal-2021-efficient,hu2023meetingbank,song-etal-2022-towards,kryscinski2021booksum}, reading comprehension~\cite{nguyen2016ms,fan2019eli5,zhong-etal-2021-qmsum,yang-etal-2023-oasum}, long-form generation~\cite{guan2021openmeva,deng2022model,roziere2023code}, and complex reasoning~\cite{wei2022chain,yao2023tree,chen2023skills}.

However, it is challenging to extend the context window from different viewpoints:
First, the predominant LLM model uses Transformer architecture~\cite{vaswani2017attention}.
Such models like BERT~\cite{devlin2018bert}, GPT~\cite{openai2023gpt4}, and T5~\cite{raffel2020exploring} employ full attention in each layer which inherently incurs quadratic time and memory complexity.
This may potentially diminish the efficiency of both the training and inference processes.
Second, attention computation over an extremely long sequence might lead to an almost even distribution, potentially causing the omission of vital information~\cite{han2023lm}.
This may further lead to the issue of being ``lost in the middle''~\cite{liu2023lost}.
Finally, the distribution of training signals for long and short sequences is imbalanced.
It is evident that longer sequences are infrequent in both plain text and instruction-tuning data.
Consequently, this rarity poses a challenge in effectively capturing long-term dependencies during the training process.

To tackle the above issues, we propose to group local-global attention layers into blocks during the training and inference phases.
This strategy enhances efficiency while yielding results comparable to those of a global attention Transformer. 
Notably, it attains equivalent performance levels with merely half the computational effort required for training.
Additionally, this approach significantly reduces memory consumption during inference by maintaining a local Key-Value (K-V) cache specifically for the local attention layers.

In Section~\ref{ssec: arhictecture}, we list the two critical components essential for a long-context model as well as the potential alternatives for consideration.
These encompass diverse attention mechanisms and methodologies for positional embedding.
Subsequently, in Section~\ref{sec: arch experiments}, we conduct a comparative analysis of these alternatives, presenting their empirical outcomes for a comprehensive evaluation.
Integrating these insights, we name our model \MODEL, drawing an analogy to the alternating black and white stripes of a zebra, which resemble the grouped local and global layers in our model's architecture.

To validate the proposed model at large scales, Section~\ref{sec: lcat} details the continuation of training the Llama-2-7B model~\cite{touvron2023llama} using long-context adaptation training through \MODEL.
This approach not only exhibits comparable performance on short-sequence benchmarks but also achieves superior perplexity results for longer sequences.
Additionally, in Section~\ref{sec: lit}, we conduct fine-tuning of \MODEL~using a combination of both short and long instruction-tuning datasets.
This is followed by a systematic evaluation of the model's performance across a range of benchmark datasets.
It demonstrates generally better performance on both long and short benchmarks compared to Llama-2-7b-chat.
To Conclude, our contribution is 3-fold:
\begin{itemize}
    \item We develop a novel architecture, referred to as \MODEL, which incorporates grouped local-global attention layers and rotary positional embedding.
    \item We conduct comprehensive experiments and detailed analyses of the \MODEL~framework across various settings, including pretraining from scratch, continuation of training, and extensive instruction tuning. The findings from these results demonstrate the advantage of \MODEL~model architecture.
    \item Additionally, we analyze the training and inference efficiency for \MODEL~and provide the pseudocode for implementation.
\end{itemize}

\section{\MODEL}
\label{sec: zebra model}
\subsection{Model Architecture Design}
\label{ssec: arhictecture}
To extend the context window for Transformer models, two critical elements must be addressed:
First, the \textit{Attention} mechanism that allows the model to efficiently focus on and process relevant parts of long sequences.
However, it is important to note that the computational of attention escalates quadratically, leading to a decrease in efficiency as the length of the sequence increases. 
Consequently, addressing this computational challenge is essential for maintaining effectiveness over longer sequences.
Second, the \textit{Positional Embedding} that imparts a structured signal indicative of the sequential order of tokens.
It is vital to employ a positional embedding that is not only robust but also exhibits strong generalization capabilities, particularly for processing long sequences.
This ensures the model's effectiveness in maintaining sequence integrity over longer spans of data.

\subsubsection{Attention}
\label{ssec: attention}
\begin{figure*}
    \centering
    \begin{subfigure}[b]{0.22\linewidth}
        \centering
        \includegraphics[width=\linewidth]{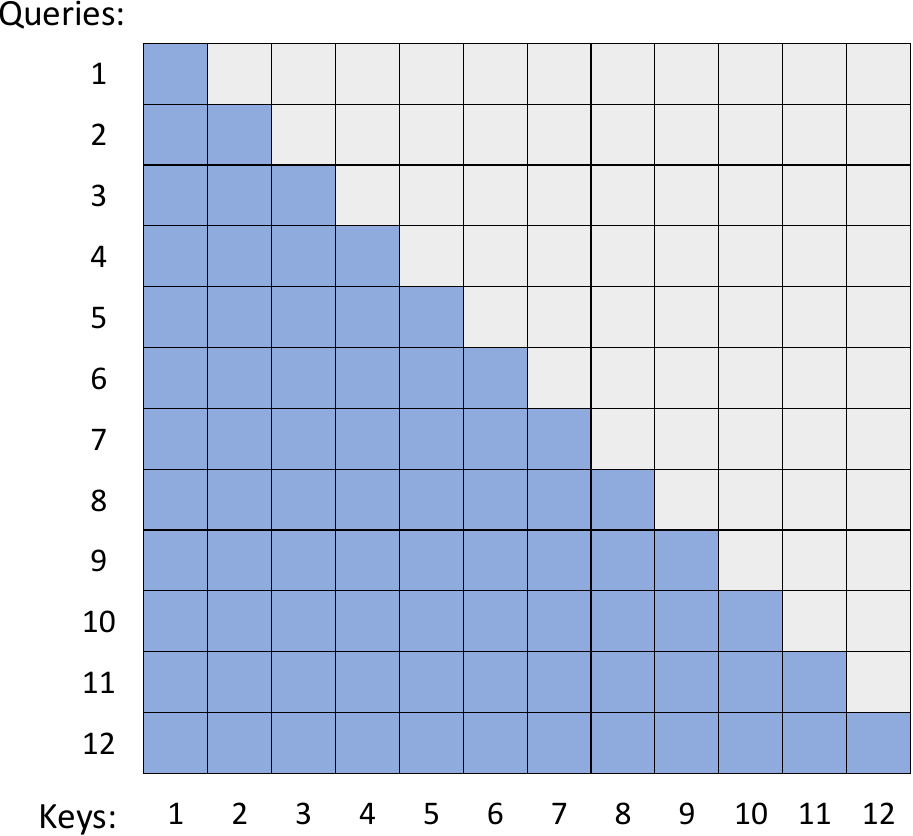}
        \caption{\small~Global Attention}    
        \label{fig: attention-global}
    \end{subfigure}
    \hfill
    \begin{subfigure}[b]{0.22\linewidth}  
        \centering 
        \includegraphics[width=\linewidth]{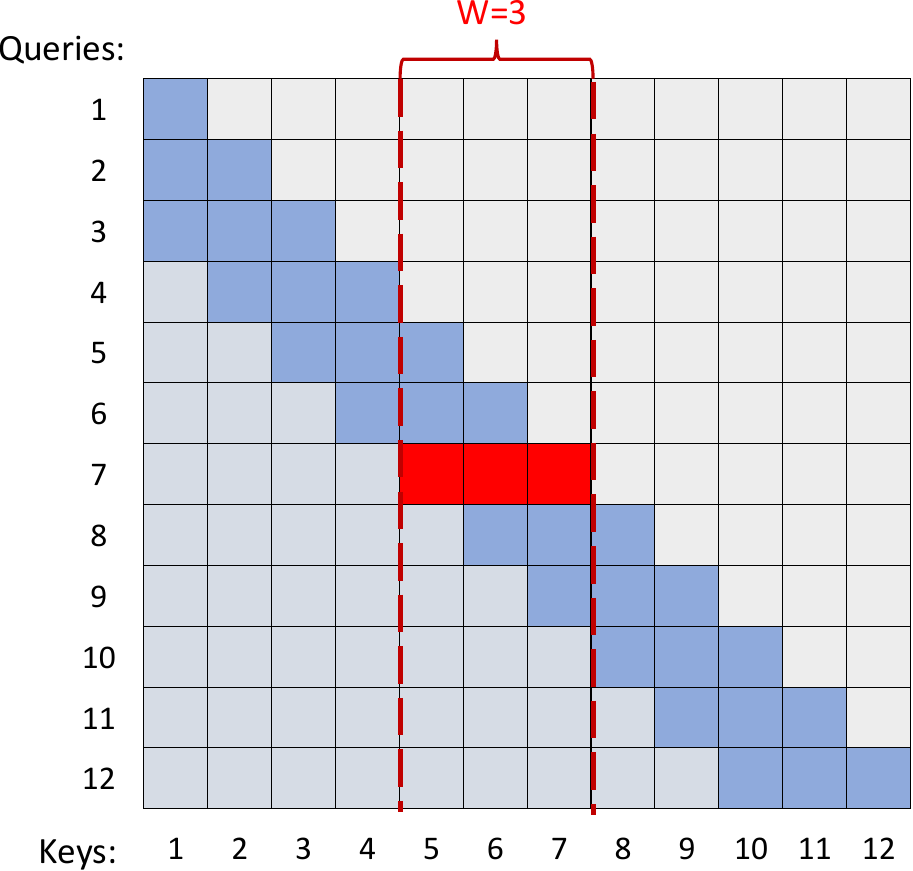}
        \caption{\small~Local Attention}
        \label{fig: attention-local}
    \end{subfigure}
    \hfill
    \begin{subfigure}[b]{0.22\linewidth}   
        \centering 
        \includegraphics[width=\linewidth]{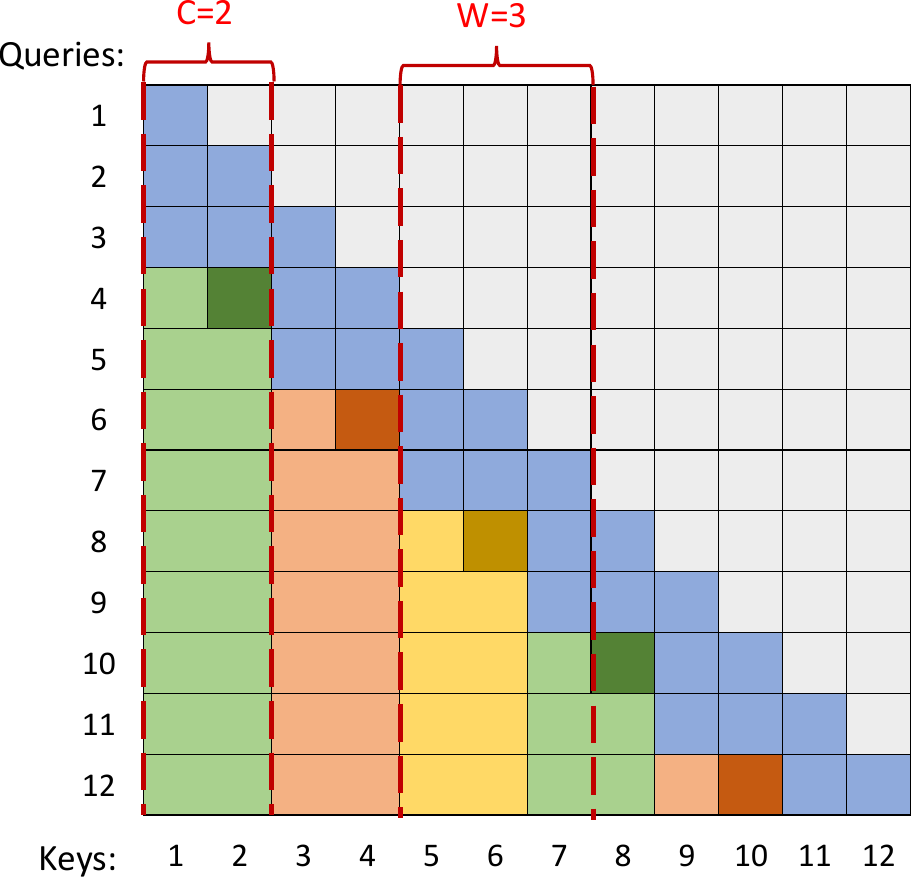}
        \caption{\small~Global-Approximation}
        \label{fig: attention-local-ga}
    \end{subfigure}
    \hfill
    \begin{subfigure}[b]{0.3\linewidth}   
        \centering 
        \includegraphics[width=\linewidth]{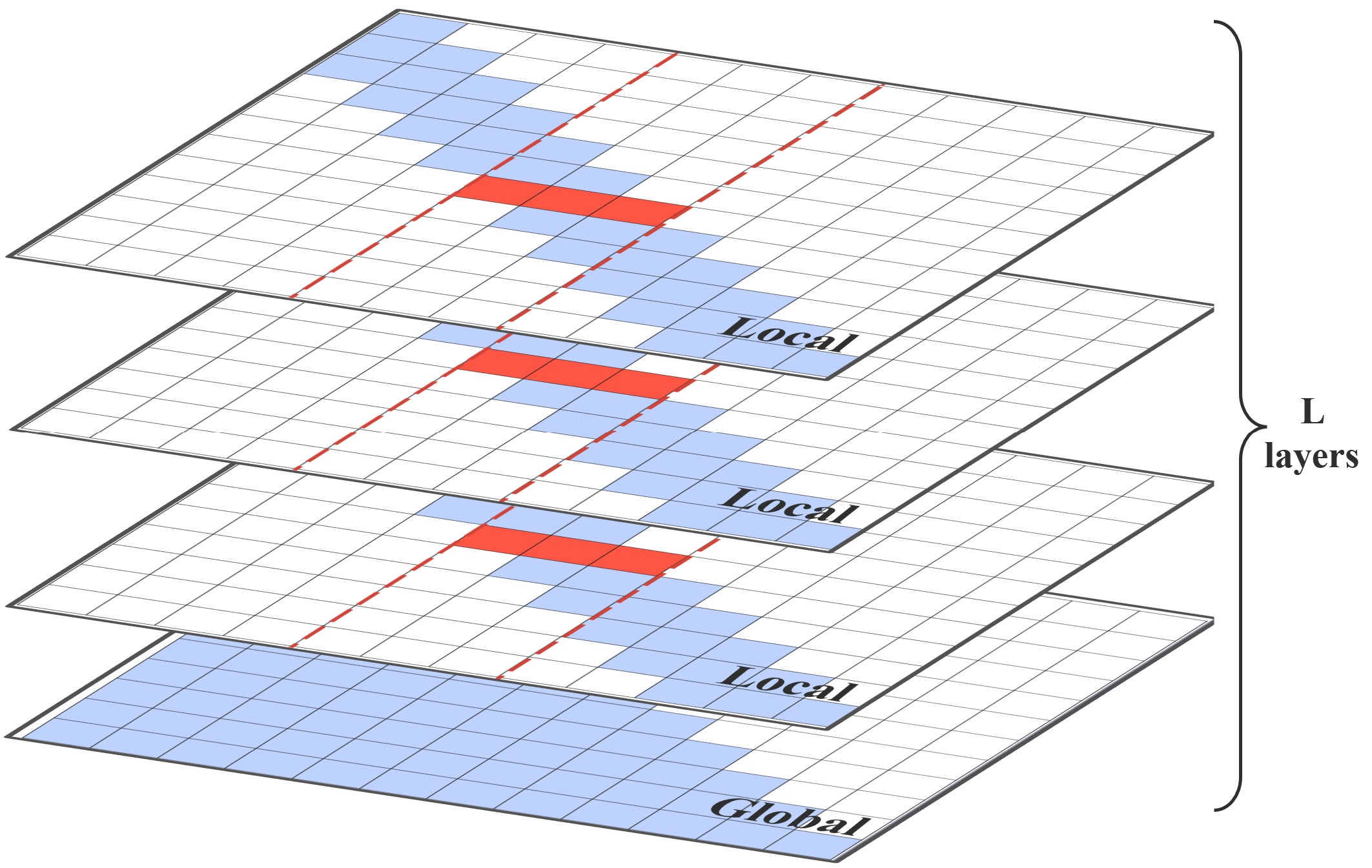}
        \caption{\small~Group Attention}
        \label{fig: attention-group}
    \end{subfigure}
    \caption
    {\small Four different attention strategies to be compared in this work. (a) Global Attention, where each token has its attention to all previous tokens and itself; (b) Local Attention, where each token only has the attention within its local window; (c) Local Attention with Global Approximation is newly introduced in this work, where each token not only has attention to its local window but also has an approximated attention from the remaining non-local chunks; (d) Group Attention is our introduced layerwise grouped local-global attention strategy, where we group $L$ layers and apply the global attention at the first layer of each group (the remaining layers use local attention).
    } 
    \label{fig: Attentions}
    \vspace{-0.1in}
\end{figure*}

In Figure~(\ref{fig: attention-global}, \ref{fig: attention-local}, \ref{fig: attention-local-ga}), we showcase three representative variants of single attention layers including global attention, local attention, and local attention with global approximations.
Additional sparse attention models like blockwise attention\cite{qiu2019blockwise}, dilated window attention~\cite{beltagy2020longformer}, stride attention~\cite{child2019generating}, Sinkhorn Attention~\cite{tay2020sparse}, transient global attention~\cite{guo2021longt5} are considered potential alternatives for basic local attention.
For the sake of clarity and focus in our research, we confine our analysis to two primary variants: local attention and local attention with global approximations.
This decision allows for a more targeted exploration of these specific attention mechanisms within our work.
Moreover, we also consider using different strategies among different layers.
In Figure~\ref{fig: attention-group}, we combine several local layers with one global attention layer as a group and stack such groups for the model.

Considering one head of the self-attention layer in a decoder-only transformer,
the query, key, and value of $i$-th position and $l$-th layer are defined as projections of the last layer hidden states $h^{(l-1)}_{i}$:

{\small
\begin{align}
    \mathbf{q}^{(l)}_{i}=&W_{q}^{T}\mathbf{h}^{(l-1)_{i}} \\
    \mathbf{k}^{(l)}_{i}=&W_{k}^{T}\mathbf{h}^{(l-1)_{i}} \\
    \mathbf{v}^{(l)}_{i}=&W_{v}^{T}\mathbf{h}^{(l-1)_{i}}
\end{align}
}%
We denote the similarity between $i$-th query and $j$-th key as:
\begin{equation}
    \small
    \textit{Sim}(i, j)=exp(\mathbf{q}_i^T \mathbf{k}_j / \sqrt{D})
    \label{eq: token similarity}
\end{equation}
where $D$ is a normalized factor usually equal to the model dimension.

\noindent \textbf{Global Attention}: It is the most common attention, where each token has attention to all the positions before it and itself:
\begin{equation}
    \small
    \alpha_{i, j}=\frac{\textit{Sim}(i, j)}{\sum_{t=0}^{i} \textit{Sim}(i, t)}
\end{equation}
where $\alpha_{i,j}$ is the attention value of $i$-th query over $j$-th key.
The context vector is then defined as a weighted sum of value vectors:
\begin{equation}
    \small
    context_{i} = \sum_{j=0}^{i}\alpha_{i,j} \mathbf{v}_{j}
\end{equation}

\noindent \textbf{Local Attention}: Each query only considers the key-value pairs within its local window.
\begin{equation}
    \small
    \alpha_{i,j} = \frac{\textit{Sim}(i, j)}{\sum_{t=min(0, i-w)}^{i} \textit{Sim}(i, t)}
    \label{eq: local-attention}
\end{equation}
where $w$ is the window size of local attention.

\noindent \textbf{Local Attention w/ Global Approximation}: Inspired by transient global attention~\cite{guo2021longt5}, we approximate the global attention output by combining multiple non-local tokens as a chunk, and take the attention over local tokens and non-local chunks.
Each non-local chunk's key-value pairs are estimated using the following equations:

{\small
\begin{align}
    \mathbf{\hat{k}}_j =& \sum_{t=(j-1) * c}^{j*c-1}\mathbf{k}_{t} + ln(c) \\
    \mathbf{\hat{v}}_j =& \sum_{t=(j-1) * c}^{j*c-1}\mathbf{v}_{t} + ln(c)
    \label{eq: chunk estimation}
\end{align}
}%
where $c$ is the chunk size, and $ln(c)$ is a compensation term for each chunk.

\noindent \textbf{Layerwise Grouped Local-Global Attention}: Instead of using identical layers for the entire network, we propose to use grouped local-global attention layers.
In figure~\ref{fig: attention-group}, we group every $L$ layer and use only one global attention layer at the first layer of each group.
We apply local attention described in Equation~(\ref{eq: local-attention}) for the remaining layers.
\begin{equation}
    \small
    h^{(l)} = \left\{
        \begin{array}{lr}
            \textit{G-Block} (h^{(l-1)}) & l \mod L == 0\\
            \textit{L-Block} (h^{(l-1)}) & \text{otherwise} \\
        \end{array}
        \right.
\end{equation}
To simplify, we use \textbf{Group Attention} to denote the layerwise grouped local-global attention.

\subsubsection{Positional Embedding}
\label{ssec: position embedding}
In the Transformer architecture, positional embeddings are commonly used to encode the sequence order information.
In this study, we incorporate three widely recognized types of positional embeddings to facilitate a comprehensive analysis.

\noindent \textbf{Absolute Positional Embedding}:
The vanilla Transformer~\cite{vaswani2017attention} advocates to use an absolute sinusoidal positional embedding:

{\small
\begin{align}
    \nonumber
    PE(pos, 2i) =& sin(pos/10000^{2i/d)}) \\
    \nonumber
    PE(pos, 2i + 1) =& cos(pos/10000^{2i/d})
\end{align}
}%
where $pos$ is the position index, $d$ is the model dimension, and $i$ is the iterative variable for different dimensions.
After the work of the vanilla Transformer, a trainable absolute positional embedding has been introduced~\cite{devlin2018bert,radford2018improving}, serving as a replacement for the fixed sinusoidal pattern.
Such positional embedding is directly added to the semantic embedding:
\begin{equation}
    \small
    \textit{EMB}(word, pos) = \textit{WE}(word) + \textit{PE}(pos)
    \vspace{-0.1in}
\end{equation}
where $word$ is the input token index, and $pos$ is the absolute position index.

Most recently, the relative positional embeddings~\cite{shaw2018self,yang2019xlnet} are introduced to eliminate the positional bias while improving the performance.
These approaches also facilitate the model's ability to extend its contextual window, a process known as position extrapolation.
Within this framework, two principal types of relative positional embeddings are taken into consideration.

\noindent \textbf{Alibi Positional Embedding}~\cite{alibi}, which applies the relative positional embedding by directly adding a bias term to the attention matrix.
\begin{equation}
    \small
    \alpha_{i,j} = \textit{Softmax}^{i}_{j}(\textit{Sim}(i, j) - (i - j) * m)
\end{equation}
where $m$ is a head-specific scalar and $(i-j)$ is the relative distance between query and key positions.
By canceling out the $-i * m$ term, we have 
\begin{equation}
    \small
    \alpha_{i,j} = \textit{Softmax}^{i}_{j}(\textit{Sim}(i, j) + j * m)
\end{equation}

\noindent \textbf{Rotary Positional Embedding}~\cite{su2023roformer} rotates the conjugate dimensions of query and key vectors, such that the relative distance is encoded during the attention calculation.

{\small
\begin{align}
    \mathbf{\widetilde{q}} =& (W_{q}^{T}\mathbf{h}_i)e^{\mathbf{i}(i\theta)} \\
    \mathbf{\widetilde{k}} =& (W_{k}^{T}\mathbf{h}_i)e^{\mathbf{i}(i\theta)}
    \vspace{-0.1in}
\end{align}
}%
where $\mathbf{i}$ denotes the imaginary unit, and $i$ is the positional index.
For each pair of conjugate dimensions, the similarity between query and key can be written as:
\begin{equation}
    \small
    \textit{Sim}(i, j) = \textit{RE}[(W_{q}^{T}\mathbf{h}_i)^{T}(W_{k}^{T}\mathbf{h}_j)e^{\textbf{i}(i-j)\theta}]
    \vspace{-0.1in}
\end{equation}
where $\textit{RE}$ takes the real value of the complex number.
The overall similarity is consequently defined as the cumulative measure of similarities across all corresponding dimensions.

\begin{figure*}[t]
    \centering
    \begin{subfigure}[b]{0.31\linewidth}
        \centering
        \includegraphics[width=\linewidth]{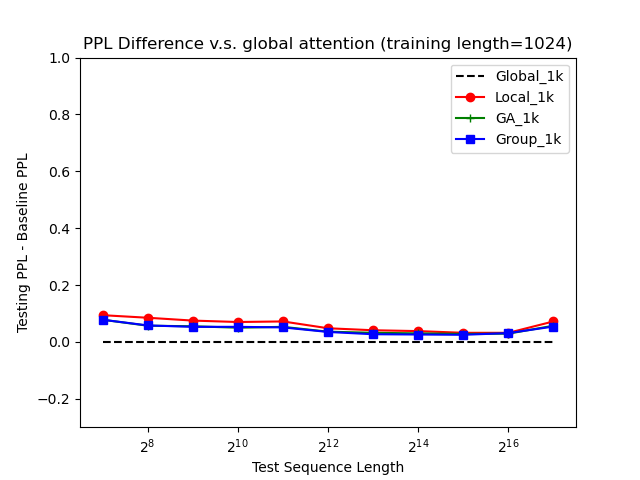}
        \label{fig: attention_1k}
    \end{subfigure}
    \hfill
    \begin{subfigure}[b]{0.31\linewidth}  
        \centering 
        \includegraphics[width=\linewidth]{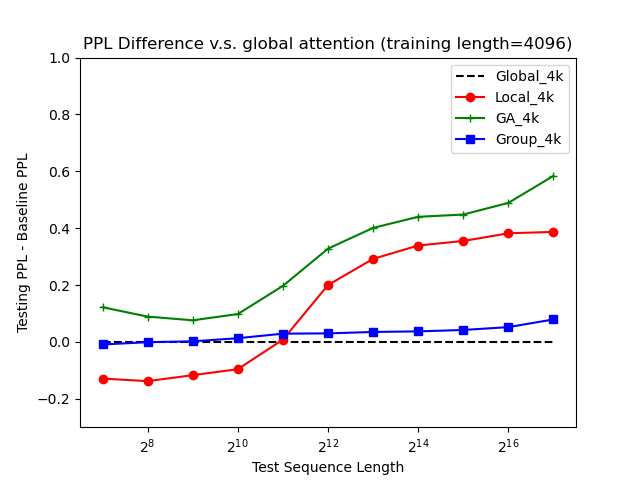}
        \label{fig: attention_4k}
    \end{subfigure}
    \hfill
    \begin{subfigure}[b]{0.31\linewidth}   
        \centering 
        \includegraphics[width=\linewidth]{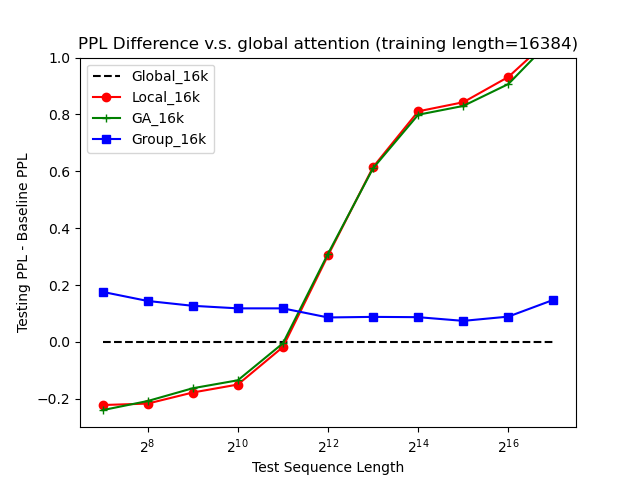}
        \label{fig: attention_16k}
    \end{subfigure}
    \vspace{-0.15in}
    \caption
    {\small The testing PPL gap between each method and the baseline system (global attention) on 1024, 4096, and 16384 training sequence length. The smaller the better. In this experiment, we split the entire testing set into different splits according to their length. Each split contains the instances within the length range of $\frac{x}{2}+1$ to $x$, except the first one (length $\leq 128$).} 
    \label{fig: attention lengths}
    \vspace{-0.15in}
\end{figure*}

\begin{figure*}[t]
    \centering
    \begin{subfigure}[b]{0.325\linewidth}
        \centering
        \includegraphics[width=\linewidth, trim = 0cm 0cm 0cm 0cm, clip]{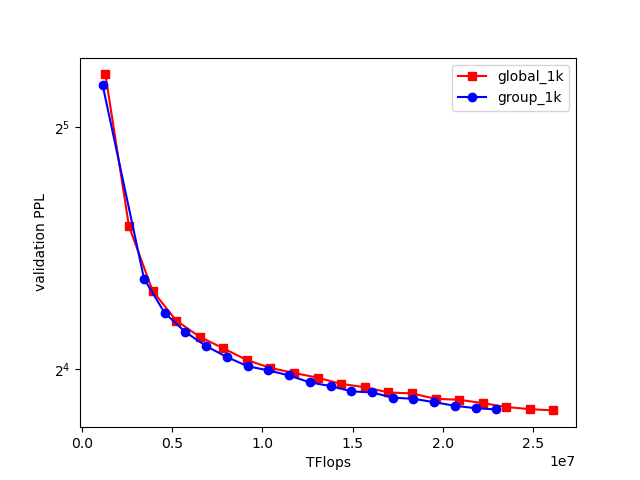}
        \label{fig: ppl_vs_tflops_1k}
    \end{subfigure}
    \hfill
    \begin{subfigure}[b]{0.325\linewidth}  
        \centering 
        \includegraphics[width=\linewidth, trim = 0cm 0cm 0cm 0cm, clip]{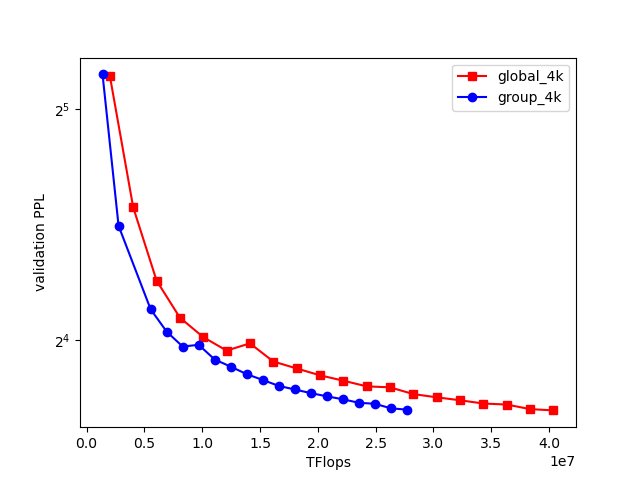}
        \label{fig: ppl_vs_tflops_4k}
    \end{subfigure}
    \hfill
    \begin{subfigure}[b]{0.325\linewidth}   
        \centering 
        \includegraphics[width=\linewidth, trim = 0cm 0cm 0cm 0cm, clip]{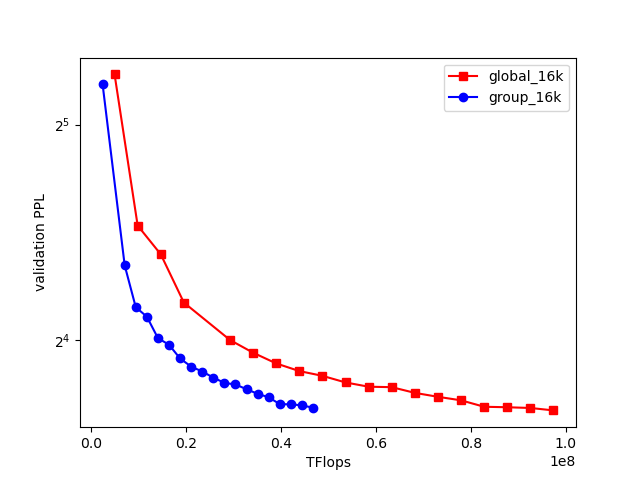}
        \label{fig: ppl_vs_tflops_16k}
    \end{subfigure}
    \vspace{-0.15in}
    \caption
    {\small The validation PPL vs TFLOPs for global attention(red) and group attention(blue) on 1024, 4096, and 16384 training sequence lengths.} 
    \label{fig: ppl_vs_tflops}
    \vspace{-0.15in}
\end{figure*}

\subsection{Experiments for Model Design}
\label{sec: arch experiments}
\begin{table}[htbp]
    \small
    \centering
    \begin{tabular}{|c|c|c|}
        \hline
        Model Size & 117M & 345M \\
        \hline
        \hline
        Num Layers & 12 & 24 \\
        Hidden Size & 768 & 1024 \\
        Num Heads & 12 & 16 \\
        K-V  Channel & 64 & 64 \\
        FF Layer Hidden Size & 3072 & 4096 \\
        \hline
    \end{tabular}
    \caption{\small Parameters of two models with different sizes.}
    \label{tab: model_size}
    \vspace{-0.15in}
\end{table}

We conduct experiments with various attention strategies and positional embedding methods as described earlier.
Two GPT models with 117M and 345M parameters as detailed in Table~\ref{tab: model_size}, are trained from scratch to assess different model architectures.
The training sequence length used for these experiments ranges from $1024$, $4096$, to $16384$.
A 10\% of the training data from the Pile dataset~\cite{gao2020pile} is utilized for model training. Its testing and validation data is used in experiments of this section for evaluation.
We employ an Adam Optimizer~\cite{kingma2014adam} for training with the beta values of $0.9$ and $0.99$.
The training process spans $20,000$ steps with a batch size of 2M tokens.
The initial learning rate is set to $1e^{-3}$ with a warm-up step of $2,000$, followed by linear decay to $1e^{-5}$.
Weight decay is set to $0.01$, and the gradient clipping is set to $1.0$.
For the local attention, a window size of $w=1,024$ is applied.
For local attention with global approximation, we employ a chunk size of $c=16$.
We group every three layers for local and global layers.
For rotary embedding~\cite{su2023roformer}, the RoPE theta is configured as $131,072$ to enhance its generalization performance on longer sequences.
All experiments are implemented using Megatron-LM\footnote{\url{https://github.com/NVIDIA/Megatron-LM}}~\cite{shoeybi2019megatron,narayanan2021efficient,korthikanti2023reducing}.

\subsubsection{Attention}
Figure~\ref{fig: attention lengths} shows the testing perplexity (PPL) difference between each attention strategy and the global attention method on the 117M model.
From the figures, we have a few observations:
First, global attention has the best overall performance;
Second, the performance gap between group attention and global attention is small but stable when the training sequence length is getting longer;
Third, as the training sequence length grows, the performance of local attention and global approximation attention drops a lot for longer sequences, though it may benefit the shorter ones.
As group attention has less computation but achieves a similar performance compared to global attention, it has a high scalability potential.

To better compare the global and group attention strategies, we take both performance and computation into consideration.
In Figure~\ref{fig: ppl_vs_tflops}, we draw the curve of the estimated TFLOPS and the validation PPL on three different training lengths with DeepSpeed FLOPS profiler\footnote{\url{https://www.deepspeed.ai/tutorials/flops-profiler/}} for the 117M model.
We observe that group attention achieves a similar performance with less computation than global attention.
When the local window is equal to the training sequence length (i.e., 1k training length in Figure~\ref{fig: ppl_vs_tflops}), the gain is negligible.
However, as the training sequence gets longer (e.g., 4k or 16k training length in Figure~\ref{fig: ppl_vs_tflops}), the gain becomes magnified.
This verifies that group attention has better scalability compared to global attention.

\subsubsection{Positional Embedding}
\label{ssec: experiments of position embeddings}
\begin{table*}[htbp]
    \small
    \centering
    \begin{tabular}{|*{13}{c|}}
        \hline
        \multicolumn{2}{|c|}{Pos. Emb.} & \multicolumn{11}{c|}{PPL on Pile Testset with Different Lengths} \\
        \hline
        \multicolumn{2}{|c|}{Min Len} & 1 & 129 & 257 & 513 & 1025 & 2049 & 4097 & 8193 & 16385 & 32789 & 65537 \\
        \multicolumn{2}{|c|}{Max Len} & 128 & 256 & 512 & 1024 & 2048 & 4096 & 8192 & 16384 & 32768 & 65536 & 131072 \\
        \hline
        \multirow{3}{*}{\rotatebox[origin=c]{90}{117M}} & Absolute & 24.62 & 20.34 & \textbf{17.00} & 17.06 & 16.42 & \textbf{11.84} & \textbf{10.02} & \textbf{8.84} & - & - & - \\
        & Alibi & \textbf{24.54} & \textbf{20.29} & 17.01 & \textbf{17.05} & \textbf{16.41} & 11.87 & 10.08 & 8.93 & \textbf{7.60} & \textbf{8.83} & \textbf{18.47} \\
        & Rotary & 24.70 & 20.37 & 17.03 & \textbf{17.05} & \textbf{16.41} & \textbf{11.84} & 10.06 & 8.92 & 7.62 & 8.86 & 18.51 \\
        \hline
        \multicolumn{2}{|c|}{$\Delta$PPL (Max - Min)} & 0.16 & 0.08 & 0.03 & 0.01 & 0.01 & 0.03 & 0.06 & 0.09 & 0.02 & 0.03 & 0.04 \\
        \hline
        \multirow{3}{*}{\rotatebox[origin=c]{90}{345M}} & Absolute & 19.41 & 16.12 & 13.57 & 13.61 & 13.19 & 9.86 & 8.37 & 7.46 & - & - & - \\
        & Alibi & 19.27 & 16.02 & 13.51 & 13.55 & 13.11 & 9.78 & 8.36 & 7.44 & 6.44 & 7.38 & 14.84 \\
        & Rotary & \textbf{19.25} & \textbf{16.01} & \textbf{13.48} & \textbf{13.51} & \textbf{13.08} & \textbf{9.74} & \textbf{8.32} & \textbf{7.42} & \textbf{6.43} & \textbf{7.37} & \textbf{14.77} \\
        \hline
        \multicolumn{2}{|c|}{$\Delta$PPL (Max - Min)} & 0.16 & 0.11 & 0.09 & 0.10 & 0.11 & 0.12 & 0.05 & 0.04 & 0.01 & 0.01 & 0.07 \\
        \hline
    \end{tabular}
    \caption{\small Perplexity with different positional embeddings and model sizes on the Pile test set. The minimum value of each model is indicated in bold. All the systems are trained with $16,384$ sequence length. In this experiment, we split the entire testing set into different splits according to their length. Each split contains the instances within the length range of its minimum length to its maximum length. $\Delta$PPL is the gap between the best-performing system and the worst-performing system of each test split.} 
    \label{tab: position embedding} 
    \vspace{-0.1in}
\end{table*}

Table~\ref{tab: position embedding} shows the perplexity results comparing different positional embeddings with the 117M and 345M models.
We find that no significant performance differences are observed among the three positional embeddings for sequence lengths within the training sequence 16,384. 
This result is in line with \cite{taylor2022galactica, kazemnejad2023impact} observation.
While the absolute positional embedding encounters challenges in extrapolating to longer sequences, both Alibi and Rotary positional embeddings demonstrate similar capabilities for sequences exceeding the training sequence length of 16,384. 
It is important to note that, in our experiments, the Alibi positional embedding requires full precision (fp32) computation to prevent position collision.
Consequently, we opt for the Rotary positional embedding in the \MODEL~model.

\subsubsection{Training Sequence length}
\label{ssec: experiments of training sequence length}
\begin{figure}[htbp]
    \centering
    \includegraphics[width=\linewidth]{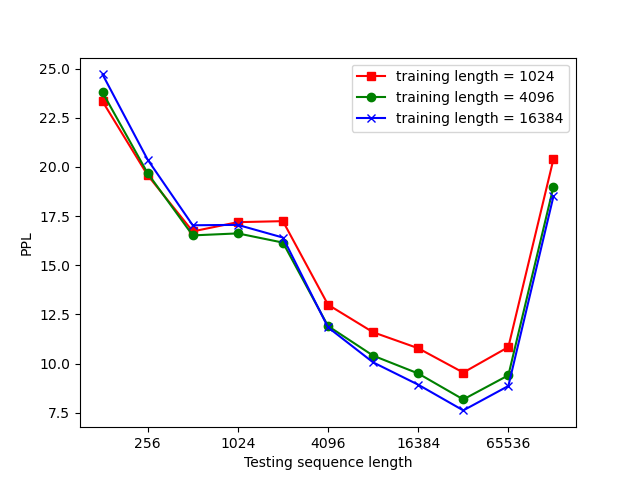}
    \caption{\small Perplexity on test sequences with 1k, 4k, and 16k training sequence lengths. In this experiment, we split the entire testing set into different splits according to their length. Each split contains the instances within the length range of $\frac{x}{2}+1$ to $x$, except the first one (length $\leq 128$).}
    \label{fig: training length}
    \vspace{-0.15in}
\end{figure}

We experiment with training sequence lengths of 1024, 4096, and 16384 with a 117M model.
The corresponding validation perplexity with the three training sequence lengths is $14.29$, $12.98$, and $12.76$, respectively.
In Figure~\ref{fig: training length}, 
we observe training with longer sequence length generally performs better than those training with shorter sequence length, especially on longer test splits.
Meanwhile, the perplexity of the model with longer training length drops a little on short test splits.
Interestingly, as we only train with 16k length, the perplexity is still going down on the 32k test split.
The results suggest that training with a longer sequence length helps with performance.

\subsection{Conclusion on Model Architecture}
\label{ssec: zebra conclusion}
Based on the experiments and analysis above, we decide to apply Rotary positional embedding and group attention strategy for extending LLM's context window.
The model is denoted as \MODEL~due to the analogous arrangement of grouped local and global attention layers, resembling the black and white color stripes on a zebra.
\section{Long Context Adaptation Training}
\label{sec: lcat}
In this section, we expand our examination of the \MODEL~architecture by applying it to a larger-scale model, utilizing Long Context Adaptation Training (LCAT).
LCAT is essential for handling large contexts.
This modification enables the model to adapt to an expanded context through \MODEL~architecture, while also providing extended signals during training. 
This approach allows the model to effectively learn and utilize long-range dependencies.
Training such a model from scratch requires a lot of computing power, so we start with an already developed model called \llama-2~\cite{touvron2023llama} as our base.
From there, we train the \MODEL~model with different volumes of data.
All our experiments are conducted with the 7B-sized model.

\subsection{Training Details}
\label{ssec: training details}
\MODEL~layers are organized into groups, each consisting of four layers ($L=4$). 
The first layer within each group functions as the global layer, while the remaining three layers serve as local layers with an attention local window size of $512$.
Each batch contains $1,572,864$ tokens.
As shown in Table~\ref{tab:zebra_pretrained_eval}, \MODEL~models are trained with $24,576$ or $32,768$ sequence lengths with different data sizes where multiple documents are packed with a $\text{BOS}$ token and an $\text{EOS}$ token.
The Adam Optimizer~\cite{kingma2014adam} is utilized, with beta values of $(0.9, 0.95)$ and an Adam epsilon of $1e^{-5}$. 
Training incorporates a cosine learning schedule with a linear warm-up of 100 steps. 
The maximum learning rate is $1e^{-4}$, and the minimum learning rate is $1e^{-5}$. 
Gradient clipping is applied with a $\num{1.0}$ threshold and weight decay is set to $\num{0.1}$.

\subsection{Data Recipes}
\label{ssec: data recipes}
\begin{table*}[htbp]
    \small
    \centering
    \begin{tabular}{|c|c|c|c|c|}
        \hline 
        \multirow{2}{*}{Domain} & \multirow{2}{*}{Data Source} & \multicolumn{3}{c|}{Data Version}\\\cline{3-5}
        & & v0 & v1 & v2 \\
        \hline
        \hline
        Book & BookCorpus2, Books3 & 100GB & 50GB & 100GB \\
        \hline
        WebPage & Pile-CC, OpenWebText2, HackerNews & 0 & 50GB & 100GB\\
        \hline
        \multirow{2}{*}{Knowledge}  & USPTO Backgrounds, PubMed Abstracts & \multirow{2}{*}{0} & \multirow{2}{*}{50GB} & \multirow{2}{*}{100GB} \\
        & Wikipedia(en), NIH ExPorter & & &\\
        \hline
        Q\&A & Stack Exchange & 0 & 20GB & 30GB\\
        \hline
        Code & Github & 0 & 20GB & 20GB\\
        \hline
        Translation & EuroParl & 0 & 10GB & 10GB\\
        \hline
        \hline
        Total & - & 100GB & 200GB & 360GB \\
        \hline
        \# Token Used & - & 15.7B & 50.3B & 100.7B \\
        \hline
    \end{tabular}
    \caption{\small Data source for each domain and various combinations of data size employed for LCAT training.} 
    \label{tab: lcat data source} 
\end{table*}

\begin{figure}[tbp]
    \centering
    \includegraphics[width=\linewidth]{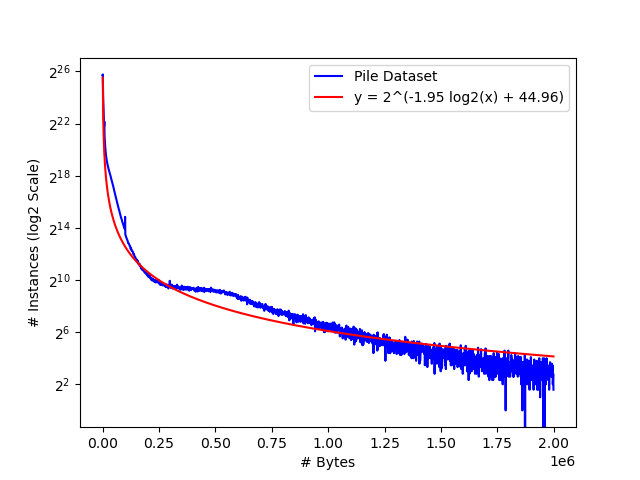}
    \caption{Sequence Length vs. Number of Instances on The Pile Dataset.
    }
    \label{fig: seq len}
\vspace{-0.15in}
\end{figure}
Figure~\ref{fig: seq len} presents an analysis of the distribution imbalance in the Pile dataset~\cite{gao2020pile}, particularly highlighting the variations due to different sequence lengths.
A notable observation is that doubling the sequence length, when measured in bytes, results in a reduction of the dataset's volume to merely one-fourth of its initial size.
In order to sufficiently capture longer text segments in our model, we have strategically integrated the LCAT data with sources that typically contain longer data, such as books.
All data is extracted from diverse sources within the Pile dataset, followed by recombination with various domains.
Table~\ref{tab: lcat data source} illustrates the detailed mapping of the original data source to the domain split.
The data within each domain split is shuffled, and the specified data quantities, as outlined in Table~\ref{tab: lcat data source}, are selected.

\subsection{Evaluation Results}
\label{ssec: lcat evaluation results}
\begin{table*}[htbp]
    \small
    \centering
    \begin{tabular}{|*{11}{c|}}
        \hline
        \multirow{2}{*}{Model Architecture} & \multicolumn{2}{c|}{Training} & \multicolumn{2}{c|}{RoPE Parameters} & \multicolumn{6}{c|}{PPL with Different Test Length}\\
        \cline{2-11}
         & Len. & Data & Scale & theta & 4096 & 8192 & 16384 & 24576 & 32768 & Average\\
        \hline
        \llama 2 & - & - & 1 & 10,000 & 7.57 & 126.53 & 1008.50 & 2,037.35 & 2814.72 & 1198.93\\
        \hline
        \multirow{4}{*}{Llama2-PI} & \multirow{4}{*}{-} & \multirow{4}{*}{-} & 4 & 10,000 & 15.27 & 15.31 & 18.35 & 64.12 & 178.97 & 58.40\\
        & & & 4 & 40,000 & 48.71 & 56.96 & 71.75 & 82.38 & 101.17 & 72.19\\
        & & & 16 & 160,000 & 195.06 & 202.61 & 212.50 & 220.79 & 240.52 & 214.30\\
        & & & 16 & 160,000 & 239.38 & 255.36 & 339.49 & 427.84 & 532.97 & 359.01\\
        \hline
        \llama 2-LCAT & 32k & v1 & 16 & 160,000 & 7.13 & 6.85 & 6.75 & 6.63 & 6.55 & 6.78\\
        \hline
        \multirow{4}{*}{\MODEL-LCAT}& 24k & v0 & 16 & 160,000 & 9.02 & 8.80 & 8.50 & 8.52 & 8.41 & 8.65\\
        & 24k & v1 & 16 & 160,000 & 7.32 & 7.06 & 6.92 & 6.85 & 6.91 & 7.02\\
        & 32k & v1 & 16 & 160,000 & 7.36 & 6.99 & 6.84 & 6.73 & 6.75 & 6.93\\
        & 32k & v2 & 16 & 160,000 & 7.14 & 6.90 & 6.71 & 6.62 & 6.57 & 6.79\\
        \hline
    \end{tabular}
    \caption{\small PPL on Gutenberg (PG-19). The data for this evaluation is from the Pile training split but excluded in LCAT training.
    } 
    \label{tab: ppl_gutenberg} 
    \vspace{-0.1in}
\end{table*}

The long context adaptation training is essential for \MODEL~to refine the initial parameters from the \llama-2 model.
Due to the architectural disparities between the \llama~model and the \MODEL~model, our objective is to achieve comparable performance with \llama-2 on both the short-context tasks and the long-context tasks after training with long sequences.
We first evaluate the pre-trained models in Table~\ref{tab: ppl_gutenberg} on Gutenberg (PG-19)~\cite{raecompressive2019} by computing the perplexity across various lengths.
Notably, \MODEL-LCAT trained with an extended context length of 32k and larger data (v2) exhibits the most favorable perplexity results, comparable to the performance of the \llama-2-LCAT model.

Additionally, we assess the performance of our models on a set of common pretraining benchmarks to ensure robust performance across standard short-context tasks~\cite{touvron2023llama,xiong2023effective}.
As shown in Table~\ref{tab:zebra_pretrained_eval}, the \llama-2-LCAT 32k model continually trained from the original \llama-2 model using the 50B-token v1 data in Table~\ref{tab: lcat data source}
results in a slight degradation in performance compared to \llama-2.
This can be attributed to the impact of long context adaptation training, as observed in previous work~\cite{chen2023extending}, which results in degradation on short-context benchmarks.
Given that global attention typically yields superior performance compared to local attention~\cite{rae-razavi-2020-transformers, beltagy2020longformer, sun-etal-2023-length}, the \llama-2-LCAT performance indicates potential performance upper limit of models with local attention.
Comparing \llama-2-LCAT to \MODEL-LCAT trained with the same amount of data (50B), the results demonstrate similarities except for MMLU.
We speculate the performance decrease in MMLU is originated from the architectural change of the model, potentially resulting in the forgetting of partial knowledge in \llama-2.
Finally, as illustrated in Table~\ref{tab:zebra_pretrained_eval}, training \MODEL~models with varying token quantities indicates that more tokens contribute to narrowing the performance gaps relative to the \llama-2-LCAT model.

\begin{table}[t]
\setlength{\tabcolsep}{3pt}
\small
\centering
\begin{footnotesize}
\begin{tabular}{|c|c|c|c|c|c|}
\hline
Model & Tks & MMLU & CS & OQA & Avg \\
\hline
\hline
\llama 2 & 2T &45.3 & 63.9 & 48.9 & 52.7 \\
\llama 2 $\textsc{Long}$ & 2T+400B & 47.8 & 64.9 & 51.0 & 54.6 \\
\llama 2-LCAT & 2T+50B & 44.4 & 61.4 & 45.6 & 50.5 \\
\hline
Zebra-LCAT & 2T+15B & 32.6 & 59.4 & 41.0 & 44.3 \\
Zebra-LCAT & 2T+50B & 39.1 & 61.2 & 46.3 & 48.9 \\
Zebra-LCAT & 2T+100B & 41.8 & 61.3 & 46.0 & 49.7 \\
\hline
\end{tabular}
\end{footnotesize}
\caption{\small Perfomance of 7B models on short-context benchmarks. The scores in the first two rows are from ~\cite{xiong2023effective}. Commonsense (CS) score as the average of PIQA~\citep{Bisk2020}, SIQA~\citep{sap-etal-2019-social}, HellaSwag~\citep{zellers-etal-2019-hellaswag}, WinoGrande~\citep{sakaguchi2019winogrande}, ARC easy and challenge~\citep{allenai:arc}, OpenBookQA~\citep{OpenBookQA2018} and CommonsenseQA~\citep{talmor-etal-2019-commonsenseqa}; OpenQA (OQA) score as the average of 5-shot performance on NaturalQuestions~\citep{Natural_Questions} and TriviaQA~\citep{2017arXivtriviaqa}.}
\label{tab:zebra_pretrained_eval}
\vspace{-0.1in}
\end{table}

\subsection{Conclusion on LCAT}
\label{ssec: lcat conclusion}

Our model exhibits comparable performance to the full attention model (\llama-2-LCAT) in both perplexity and downstream tasks. 
This equivalence is achieved while ensuring faster training and higher GPU throughput. 
Based on the experimental findings, we choose \MODEL-LCAT trained with a 32k context window and 100B tokens for subsequent instruction tuning (a.k.a. supervised fine-tuning) in long context window, as outlined in Section~\ref{sec: lit}.
\section{Long Instruction Tuning}
\label{sec: lit}

The Long Context Adaptation Training (LCAT), as discussed in Section~\ref{sec: lcat}, facilitates the adaptation of the \MODEL~model architecture from its \llama-2 foundation model architecture.
This adaptation involves the incorporation of grouped local-global layers and position interpolation by training the model with the pre-training learning objectives in long context.
Furthermore, we hypothesize that LCAT contributes to enhancing the model by incorporating additional knowledge and capabilities that can only be acquired through training in longer context windows, as suggested by prior research~\cite{xiong2023effective}.

To align the \MODEL-LCAT model for comprehensive open-domain language understanding and generation tasks based on its acquired knowledge, we conduct supervised fine-tuning on the \MODEL-LCAT model with Long Instruction Tuning (LIT). 
The objective is to empower the model to proficiently handle both short and long-context tasks, adhering to the specifications outlined in human instructions.

\subsection{Instruction Tuning Data Recipe}
To align the model for both short and long-context tasks, our instruction tuning (IT) data comprises instances of both types. 
Table~\ref{tab:zebra_lit_data} shows the statistics on the average token number per instance.
The \llama~tokenizer is utilized for tokenizing the instances to calculate the token numbers.

\begin{table}[ht]
\setlength{\tabcolsep}{3pt}
\small
\centering
\begin{footnotesize}
\begin{tabular}{|c|c|c|c|c|c|c|}
\hline
 & \# of Instances & Avg. Tokens \# / Instance \\
\hline
\hline
Short IT Data & 344,818 & 208.3 \\
Long IT Data  & 108,963 & 6113.5 \\
\hline
\textit{Overall} & 453,781 & 1626.3 \\
\hline
\end{tabular}
\end{footnotesize}
\caption{\small Statistics of our instruction tuning (IT) data for Zebra LIT. Our short instruction tuning data (Short IT Data) contains more instances than our long instruction tuning data (Long IT Data) but Long IT Data has a significantly larger average token number per instance.}
\label{tab:zebra_lit_data}
\vspace{-0.1in}
\end{table}

\subsubsection{Short Instruction Tuning Data}
Our short instruction tuning data primarily incorporates publicly available English instruction tuning datasets, including LIMA~\cite{Zhou2023LIMALI}, Alpaca~\cite{alpaca}, and ShareGPT~\cite{shreGPT}. 
Additionally, we introduce samples from hh-rlhf~\cite{ganguli2022red} with ``selected'' responses for multi-turn dialogue. 
To customize the profile of our Zebra assistant, our dataset further encompasses carefully curated short instances.

\subsubsection{Long Instruction Tuning Data}
For long instruction tuning, we address two cases: First case, where the user provides a lengthy document and instructs the system to process the substantial information and complete the specified task succinctly; Second case, where the input is relatively short, but the desired output needs to be more extensive. 
The former case encompass tasks of summarization, question-answering (QA), and machine reading comprehension (MRC), while the latter involves writing tasks. 
In writing tasks, the user provides key information, major characters, or an outline, tasking the AI assistant with composing a natural language document, such as a letter or a chapter in a novel.

To systematically empower LLMs for long tasks in both scenarios, we carefully curate high-quality instruction tuning data for three specific tasks: summarization, long-MRC, and writing.

\noindent \textbf{Summarization}: We select 3,000 news reports from CNN / Daily Mail~\cite{see-etal-2017-get} and 2,000 multi-turn long dialogue transcripts from MediaSum~\cite{zhu2021mediasum} as documents to summarize. Given an document, we randomly choose one out of ten predefined prompts to instruct GPT-4~\cite{openai2023gpt4} to generate a summary. Our long instruction tuning data for summarization hence consists of $5,000$ instances with a document and the randomly selected summarization prompt as input, and the GPT-4 generated summary as output.

\noindent \textbf{Long-MRC}: We create synthetic long-MRC data utilizing the long documents from Guttenberg PG-19 corpus~\cite{raecompressive2019}. Given a long document (e.g., a chapter or a chapter chunk) in the corpus, we first divide the long document into text segments in approximately even length. For each text segment, we prompt GPT-4 to write one question-answer pair grounding on the information from the text segment. Each long-MRC instruction tuning instance then consists of the long document and the generated question as input $x$, and the generated answer as output $y$. We collect in total of 1,245 such instances.

\noindent \textbf{Writing}: We further utilize texts from Guttenberg PG-19 corpus to generate the data for writing. Given a document (e.g., a chapter or a chapter chunk) in the corpus, we prompt ChatGPT~\cite{openai2022ChatGPT} to extract its key elements including ``central idea'', ``outline'', ``keywords'', ``time'', ``place'', ``key figures'', ``cause'', ``process'', and ``result''. We use predefined rules to randomly select a subset of these elements, dynamically fill them into the instruction template, and hence use the completed instruction containing selected key elements as our input.  The original document is then the corresponding output. We collect 328 such instances.

Besides our curated data for tasks of summarization, long-MRC, and writing, we further incorporate 102k training instances randomly selected from public datasets including BigPatent~\cite{sharma-etal-2019-bigpatent}, GovReport~\cite{huang-etal-2021-efficient}, GSM8k~\cite{cobbe2021gsm8k}, CUAD~\cite{hendrycks2021cuad}, MultiDoc2Dial~\cite{Feng_2021}, Multi-News~\cite{fabbri-etal-2019-multi}, Natural Question~\cite{Natural_Questions}, Musique~\cite{trivedi-etal-2022-musique}, NarrativeQA~\cite{kocisky-etal-2018-narrativeqa}, Qasper~\cite{dasigi-etal-2021-dataset}, QMSum~\cite{zhong-etal-2021-qmsum}, QuALITY~\cite{pang-etal-2022-quality}, SPACE~\cite{angelidis-etal-2021-extractive}, SQuALITY~\cite{wang-etal-2022-squality}, SummScreen~\cite{chen-etal-2022-summscreen}, and TOEFL-QA~\cite{tseng2016machine,chung-etal-2018-supervised}. 
These datasets cover traditional long tasks like summarization, QA, and multi-choice QA.

\subsection{LIT Training Details}
The \MODEL-LIT training uses the \MODEL-LCAT model as backbone and thus inherits \MODEL-LCAT's model structure including the grouped attention setting. \MODEL-LIT is trained with $l_s=16,384$ sequence length. We pack training instances with less than $l_s$ tokens into the $l_s$-token sequences. The padding tokens are added to the right. For long instruction instances with more than $l_s$ tokens, we truncate only its input document but keep all the instruction text. 
The batch size is $2,097,152$ tokens per batch. 
We use Adam Optimizer with beta values of $(0.9, 0.95)$ and an Adam epsilon of $1e^{-8}$. 
Training incorporates a cosine learning schedule with a linear warm-up of 32 steps. 
The maximum learning rate is $2e^{-5}$, and the minimum learning rate is $6e^{-6}$. 
We conduct 4-epoch training with our instruction tuning data. In LIT, we calculate the loss on the output tokens $y$ only.
The rest hyper-parameters are the same as LCAT training.

\subsection{Evaluation Results}
\label{ssec: long-eval-results}
\begin{table*}[ht]
\setlength{\tabcolsep}{4pt}
\small
\centering
\begin{footnotesize}
\begin{tabular}{|c|c|c|c|c|c|c|}
\hline
Model (7B) & cot/gsm8k & cot/math  & cot/bbh & cot/mmlu & human-eval-plus & \textit{Avg.} \\
\hline
\llama 2-chat & 25.0 & 3.8 & 30.7 & 40.7 & 9.8 & 22.0 \\
\hline
Zebra-LIT & 30.4 & 2.2 & 33.6 & 38.8 & 15.2 & 24.0 \\
\hline
\end{tabular}
\end{footnotesize}
\caption{\small Performance of instruction tuned 7B models on short-context benchmarks. In this evaluation, we follow the setting of FastEval~\cite{fastEval}, which focuses mostly on the zero-shot setting.}
\label{tab:zebra_lit_eval_short}
\vspace{-0.1in}
\end{table*}
\begin{table*}[ht]
\setlength{\tabcolsep}{4pt}
\small
\centering
\begin{footnotesize}
\begin{tabular}{|c|ccc|ccccc|c|}
\hline
\multirow{2}{*}{Model (7B)} & \multicolumn{3}{c|}{Summarization} & \multicolumn{5}{c|}{QA} & \\
     & GR & SS & QM & SQAL & Qspr & Nrtv & QALT & MuSQ & \textit{Avg.} \\
\hline
\llama 2-chat & 11.1 & 12.7 & 14.2 & 19.0 & 14.8 & 11.0 & 38.1 & 6.7 & 15.9 \\
\hline
Zebra-LIT & 13.6 & 5.3 & 13.6 & 17.3 & 21.4 & 22.6 & 33.3 & 12.0 & 17.4 \\
\hline
\end{tabular}
\end{footnotesize}
\caption{\small Performance of instruction tuned 7B models on long-context benchmarks, validation set on ZeroScrolls.}
\label{tab:zebra_lit_eval_long}
\vspace{-0.1in}
\end{table*}

We assess the performance of our instruction-tuned model on short and long-context benchmarks, as presented in Table~\ref{tab:zebra_lit_eval_short} and~\ref{tab:zebra_lit_eval_long}, respectively. 
Fine-tuning models on long instruction data reveals some degradation in performance on short-context benchmarks, such as MMLU, as depicted in Table~\ref{tab:zebra_lit_eval_short}. 
However, the model consistently outperforms the \llama-2-chat model overall.
Additionally, we evaluate the model on long-context benchmarks~\cite{shaham2023zeroscrolls}. 
The results, as shown in Table~\ref{tab:zebra_lit_eval_long}, indicate that our model performs comparably or surpasses \llama 2-chat in QA datasets but falls short in datasets employing Rouge~\cite{lin-2004-rouge} evaluation metrics. 
It is noteworthy that the long-context evaluation of LLMs poses a non-trivial challenge, as highlighted in~\cite{xiong2023effective}.
Automatic metrics, such as Rouge, employed in benchmarks, only consider $n$-gram overlaps with a reference, which may not consistently align with human preferences.
We anticipate that incorporating more substantial and diverse fine-tuning data can significantly contribute to improved model performance.
Overall, our instruction-tuned model demonstrates better performance on both short and long-context benchmarks, affirming the effectiveness of the architecture.

\section{Discussion}
\subsection{Scalability}
\label{ssec: scalability}

\begin{figure}[htbp]
    \centering
    \includegraphics[width=\linewidth]{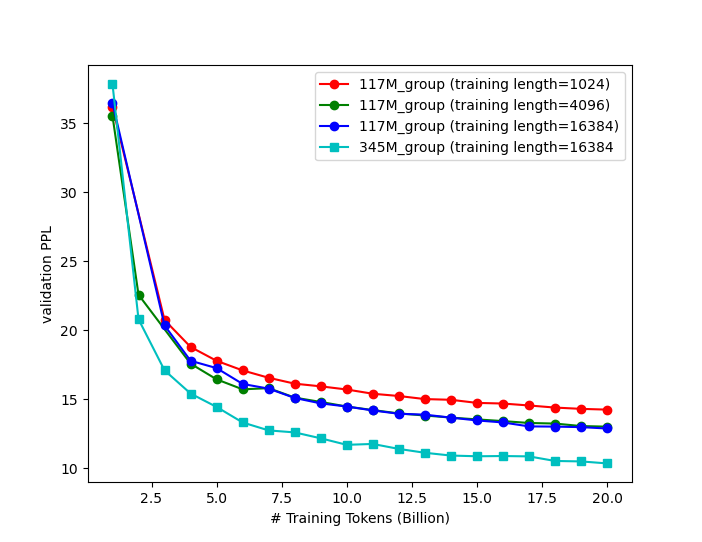}
    \caption{\small Perplexity (PPL) on Pile validation set vs number of training tokens for different model sizes and training sequence lengths.
    }
    \label{fig: scale}
\vspace{-0.1in}
\end{figure}

Due to resource constraints, our scalability experiments are limited to models with 117M and 345M parameters.
As illustrated in Figure~\ref{fig: scale}, employing a larger context window enhances model performance for the 117M model.
However, a notable observation is the difficulty in differentiating the performance curves when the sequence length is increased from $4,096$ to $16,384$ tokens.
We attribute this marginal difference to two primary factors.
Firstly, the amount of training data, capped at $20$ billion tokens, is likely insufficient to manifest a significant difference.
Secondly, both the training and validation datasets predominantly comprise shorter sequences, thereby inherently constraining the potential improvement in PPL.
Despite these limitations, we anticipate that the benefits of the increased model size will become more apparent with larger and more varied datasets.
Evidently, as the model size increases from 117M to 345M, the performance gains significantly, notably from $12.87$ to $10.34$.

\subsection{Efficiency}
\label{ssec: efficiency}
\begin{table}[htbp]
    \tiny
    \centering
    \begin{tabular}{|c|c|c|c|}
        \hline
        \multirow{2}{*}{Method} & \multicolumn{3}{c|}{$\mathcal{O}$ Complexity} \\\cline{2-4}
          & Attention & Other & Total \\
         \hline 
         Global & $DN^2$ & \multirow{4}{*}{$D^2N$} & $DN^2 + D^2N$ \\
         Local & $DWN$ &  & $D(D+W)N$ \\
         GA & $\frac{D}{C^2}N^2 + DWN$ &  & $\frac{D}{C^2}N^2 + D(D+W)N$ \\
         Group & $\frac{D}{L}N^2 + DWN$ &  & $\frac{D}{L}N^2 + D(D+W)N$ \\
         \hline
    \end{tabular}
    \caption{\small Complexities of attention and other operations.}
    \label{tab: attention complexity}
    \vspace{-0.1in}
\end{table}
\noindent \textbf{Training Efficiency}:
In Table~\ref{tab: attention complexity}, we present a detailed comparison of computational complexities for different attention operations, as well as other relevant operations, across the four distinct attention strategies outlined in Section~\ref{ssec: attention}.
This comparison indicates that with an increase in sequence length, the computational demand of global attention exhibits a quadratic growth pattern. 
On the other hand, both the Global Approximation and Group Attention strategies, though also following a quadratic growth trajectory, do so with significantly lower coefficients.
Notably, Local Attention demonstrates the best computational efficiency, requiring the least amount of resources as the sequence length extends.

\noindent \textbf{Inference Efficiency}:
Because of the implementation of local attention in the majority of Transformer layers, \MODEL~does not require retaining all Key-Value (K-V) pairs in the cache.
Consequently, this approach significantly reduces the GPU memory requirements during inference, thereby potentially increasing the inference throughput. 

The detailed pseudo-code for \MODEL~training and inference can be found in Appendix~\ref{app: implementation}.
\section{Related Work}
\label{sec: related works}

\noindent \textbf{Attention}.
The Transformer architecture has a self-attention component with $O(N^{2})$ computation complexity.
Numerous studies have been proposed to enhance the time and memory efficiency of Transformer models. 
One approach involves leveraging sparse attention patterns, enabling the transformation of full quadratic attention computations to $O(N\log N)$ or linear complexity.
Our work falls within this method by grouping sparse and full attention patterns.
Methods such as Sinkhorn~\cite{10.5555/3524938.3525813}, Longformer~\cite{beltagy2020longformer}, ETC~\cite{ainslie-etal-2020-etc}, and BigBird~\cite{zaheer2020bigbird} have been introduced to incorporate both sparse and full attention mechanisms.
Another set of approaches involves utilizing the low-rank approximation of the attention matrix.
This includes methods such as Linformer~\cite{wang2020linformer}, Performer~\cite{choromanski2022rethinking}, and Random Feature Attention~\cite{peng2021random}.

\noindent \textbf{Positional Embedding}.
In Transformer models, positional embeddings can be primarily categorized into two types: absolute and relative.
Earlier versions of Transformers utilize absolute positional encoding. For instance, the vanilla Transformer~\cite{vaswani2017attention} model adds sinusoidal positional embeddings to the word embeddings, whereas GPT~\cite{radford2018improving} and BERT~\cite{devlin2018bert} introduce learnable positional embeddings.
Currently, it has become more common to use relative positional embedding. For instance, Transformer-XL~\cite{dai-etal-2019-transformer} and T5~\cite{raffel2020exploring} adds learnable attention logit bias into attention layers.
Alibi~\cite{alibi} biases attention scores based on the distance between key and query elements.
RoPE~\cite{su2023roformer} multiplies the keys and queries of every attention layer by sinusoidal embeddings.
The Alibi and RoPE methods are further improved through the incorporation of an additional bias term~\cite{sun2022lengthextrapolatable,chi-etal-2023-dissecting}.

\noindent \textbf{LLM}.
In the early stages, open-source large language models such as OPT~\cite{zhang2022opt}, BLOOM~\cite{workshop2023bloom}, and Llama-1~\cite{touvron2023llama} have a context window length limited to 2048 tokens.
Later, a smaller open-source model Starcoder~\cite{li2023starcoder}, with 15B parameters, manage to extend its context window length to 8000 by optimizing computational efficiency with Multi-Query Attention~\cite{shazeer2019fast} and FlashAttention~\cite{dao2022flashattention}.
Following this, \llama-2~\cite{touvron2023llama}, another open-source model, expands its default length to 4096 tokens.
The open-source community then discover that by interpolating Rotary Positional 
Embeddings\footnote{\url{https://www.reddit.com/r/LocalLLaMA/comments/14fgjqj/a_simple_way_to_extending_context_to_8k}}, the context window length could be further extended to 8192 tokens.
Subsequently, \citep{chen2023extending} expand and validate this approach, known as Position Interpolation, to further enhance the window length capability.
\llama-2 undergoes extended training with long-context continual pertaining, extending up to 32,768 tokens, the positional interpolation method, and FlashAttention~\cite{dao2022flashattention} showing its enhanced effectiveness~\cite{xiong2023effective}.
Similar approaches are employed to extend the context length~\cite{peng2023yarn,du2022glm,longchat2023} by fine-tuning pretrained models with long documents.
LongLoRA~\cite{chen2023longlora} adopts a fine-tuning approach with shifted local attention for more efficient training to further extend context length.
As of December 2023, closed-source large language models have significantly expanded their context window capabilities, reaching scales of up to 100,000 tokens. 
For instance, \texttt{GPT-4-Turbo}\footnote{\url{https://platform.openai.com/docs/models/gpt-4-and-gpt-4-turbo}} supports a context window of 128,000 tokens, while \texttt{Claude-2.1}\footnote{\url{https://docs.anthropic.com/claude/reference/selecting-a-model}} supports up to 200,000 tokens.
The commercialization of these closed-source models is heavily reliant on long context understanding.
For example, it allows users to upload long text files for queries or engage in extended dialogues with extensive historical records.

\noindent \textbf{Long Evaluation}.
The majority of benchmarks for evaluating large language models are primarily focused on tasks involving short context.
However, the evaluation dimensions for long context and short context may differ significantly.
For example, ~\citep{liu2023lost} develop a multi-document question answering task, which demonstrates the phenomenon of being \textit{lost in the middle}, a challenge not present in short context.
The foundation for comprehensive evaluation of long context understanding is currently still underdeveloped.
Recently, there have been efforts to develop benchmarks specifically for long context analysis, such as~\cite{shaham2023zeroscrolls,kwan2023m4le,dong2023bamboo,bai2023longbench,an2023leval}.
\section{Conclusion}
\label{sec: conclusion}
In this work, we introduce \MODEL, a novel architecture designed to enhance the capabilities of Large Language Models (LLMs) in processing and interpreting long text sequences.
Through the innovative use of grouped local-global attention layers and rotary positional embedding, \MODEL~addresses critical challenges associated with extending the context window in LLMs.
Our extensive experiments and analyses demonstrate that \MODEL~not only maintains comparable performance on short-sequence benchmarks but also excels in handling longer sequences, as evidenced by its superior long benchmark performances and perplexity results on Gutenberg (PG-19).
This indicates that our approach effectively balances the need for efficiency with the demand for high performance in long-context scenarios.
The grouped local-global attention mechanism, in particular, proves to be a crucial component in achieving this balance, offering a significant reduction in computational and memory requirements while maintaining, and in some cases enhancing model performance.
Moreover, the application of \MODEL~among diverse short and long downstream tasks showcases its versatility and robustness across various NLP tasks.

In conclusion, \MODEL~represents a significant step forward in the realm of long-context language processing.
Its ability to efficiently handle extensive text sequences without compromising on performance opens up new possibilities for the application of LLMs in a variety of complex and information-rich environments.
We believe that \MODEL~sets a new standard for long-context modeling and will inspire further innovation in the field.

\section*{Limitations}
\label{sec: limitations}

While our work introduces a novel model architecture and exhibits promising accuracy and efficiency, it is not without limitations.

Due to computation resource constraints, we have not yet evaluated the model architecture with a parameter size larger than 7B. Though a larger model typically brings stronger performance, it is still valuable to further verify with \MODEL.

Moreover, our current evaluation, especially for long-context alignment tasks, largely relies on automatic metrics like Rouge and F-1 employed by public benchmarks. Such metrics evaluating $n$-gram overlapping with a reference have been under debate before the era of LLMs. We anticipate a comprehensive evaluation strategy for long-context alignment to be proposed in the future.


\bibliography{custom}
\bibliographystyle{acl_natbib}

\appendix

\onecolumn
\section{Appendix}
\label{sec: appendix}

\subsection{Group Attention Implementation}
\label{app: implementation}
As detailed in Section~\ref{sec: arch experiments}, our architecture employs a layerwise grouped local-global attention approach, segregating the application of local and global attention across different layers.
The methodology for local attention is outlined in Algorithm~\ref{alg: local attention}.
In contrast, the global attention mechanism adheres to the conventional multi-head attention paradigm found in transformers.
It is important to note that during inference when leveraging the Key-Value (K-V) cache, the process for local attention layers deviates from Algorithm~\ref{alg: local attention}.
Instead, we exclusively implement the standard global attention framework while maintaining the latest $w$ K-V states.
This approach streamlines the attention process while ensuring efficiency and effectiveness in inference.

\SetKwComment{Comment}{/* }{ */}
\begin{algorithm}
\small
\caption{Local Attention}\label{alg: local attention}
\KwData{$Q,K,V \in \mathcal{R}^{bsz \times len \times n\_heads \times dim\_per\_head}$}
\KwData{$M \in [0, 1]^{bsz \times n\_heads \times len \times len}$ is attention mask}
\KwData{$w$ is the window size} 
\KwData{$\lambda$ is the normalizing factor}
\tcp{padding the sequence length to multiple of window size}
\tcp{after padding shape:}
\tcp{$bsz \times padded\_len \times n\_heads \times dim\_per\_head$}
$Q \gets\textit{pad\_to\_multiple}(Q, w)$\;
$K \gets\textit{pad\_to\_multiple}(K, w)$\;
$V \gets\textit{pad\_to\_multiple}(V, w)$\;

\tcp{split $Q, K, V$ into blocks}
\tcp{after spliting shape:}
\tcp{$bsz \times n\_blocks \times w \times n\_heads \times dim\_per\_head$}
$Q_{local} \gets \textit{split\_into\_blocks}(Q)$\;
$K_{local} \gets \textit{split\_into\_blocks}(K)$\;
$V_{local} \gets \textit{split\_into\_blocks}(V)$\;

\tcp{for $K$ and $V$ merge each block and the blocks before it}
\tcp{after merging shape:}
\tcp{$bsz \times n\_blocks \times 2 * w \times n\_heads \times dim\_per\_head$}
$K_{local} \gets \textit{concatenate\_2\_blocks}(K)$\;
$V_{local} \gets \textit{concatenate\_2\_blocks}(V)$\;

\tcp{calculate attention score}
\tcp{the attention score shape:}
\tcp{$bsz \times n\_heads \times n\_blocks \times w \times 2*w$}
$\textit{attn\_score} \gets $ \;
$\quad \textit{torch.einsum}('...qhd,...khd->...hqk', Q_{local}, K_{local}) $ \;
$\quad .\textit{transpose}(1, 2)$ \;

\tcp{multiply with the normalizing factor}
$\textit{attn\_score} \gets \lambda * \textit{attn\_score}$ \;

\tcp{Extract the attention from the original attention mask}
$M_{original} \gets \textit{extract\_original\_blocks}(M, w)$ \;
\tcp{Prepare the attention for a single block ($w \times 2*w$}
$M_{block} \gets \textit{create\_blockwise\_mask}(w)$\;
\tcp{Taking both attention mask into consideration}
$M_{overall} \gets M_{original} \cap M_{block}$

\tcp{merge the sequence length dim}
\tcp{after merging shape:}
\tcp{$bsz \times n\_heads \times padded\_len \times 2*w$}
$\textit{new\_shape} \gets (bsz, n\_heads, padded\_len, 2*w)$ \;
$\textit{attn\_score} \gets \textit{attn\_score}.\textit{reshape}(\textit{new\_shape})$ \;
$M_{overall} \gets M_{overall}.\textit{reshape}(\textit{new\_shape})$ \;

\tcp{softmax}
$\textit{attn\_prob} \gets \textit{softmax}(\textit{attn\_score}, M_{overall})$ \;

\tcp{reshape back in block format}
\tcp{shape: $bsz \times n\_blocks \times w \times n\_heads \times 2 * w$}
$\textit{new\_shape} \gets (bsz, n\_heads, n\_blocks, w, 2*w)$ \;
$\textit{attn\_prob} \gets \textit{attn\_prob}.\textit{reshape}(\textit{new\_shape}).\textit{transpose}(1, 2)$ \;

\tcp{get context vector}
\tcp{shape: $bsz \times n\_blocks \times w \times n\_heads \times dim\_per\_head$}
$\textit{attn\_outputs} \gets$ \;
$\quad \textit{torch.einsum}('...hqd,...khd->...qhd', \textit{attn\_prob}, V_{local}) $ \;

\tcp{reshape to output format}
$\textit{new\_shape} \gets (bsz, padded\_len, n\_heads, dim\_per\_head)$ \;
\tcp{Don't forget to remove the padding ones}
$\textit{attn\_outputs} \gets \textit{attn\_outputs}.\textit{reshape}(new\_shape)[:, seq\_len, :, :]$ \;
\KwResult{$attn\_outputs$}
\end{algorithm}

\subsection{Case Study}
\label{app: case_study}
As discussed in Section~\ref{ssec: long-eval-results}, evaluating long-context tasks presents challenges. 
Firstly, common evaluation metrics may misrepresent response quality, particularly in summarization datasets where the reference summary may not align with diverse reader interests or knowledge levels. 
Furthermore, the limited diversity in long-context tasks may result in inadequate training for instruction tuning, often stemming from summarization, QA, or information extraction datasets with long documents. 
A more varied set of tasks could potentially enhance the capability of instruction-tuned models. 
Lastly, as demonstrated in the subsequent section, instances are frequently constructed with inaccurate gold labels or information not covered in given documents, intensifying the difficulty of model evaluation and potentially yielding erroneous results.
We present exemplars and results employing \MODEL, \llama, and ChatGPT-3.5 on ZeroScrolls~\cite{shaham2023zeroscrolls}.
Due to the limited space and long documents, summarization examples are not presented.

\newpage
\begin{example}{NarrativeQA}

\noindent\textbf{Instruction}
You are given a story, which can be either a novel or a movie script, and a question. Answer the question as concisely as you can, using a single phrase if possible. Do not provide any explanation.

\noindent\textbf{Document}
[...] INTRODUCTION. The Crito seems intended to exhibit the character of Socrates in one light only, not as the philosopher, fulfilling a divine mission and trusting in the will of heaven, but simply as the good citizen, who having been unjustly condemned is willing to give up his life in obedience to the laws of the state... \hl{The days of Socrates are drawing to a close;} the fatal ship has been seen off Sunium, as he is informed by his aged friend and contemporary Crito, who visits him before the dawn has broken; he himself has been warned in a dream that on the third day he must depart. Time is precious, and \hl{Crito has come early in order to gain his consent to a plan of escape.} This can be easily accomplished by his friends, who will incur no danger in making the attempt to save him, but will be disgraced for ever if they allow him to perish. He should think of his duty to his children, and not play into the hands of his enemies. Money is already provided by Crito as well as by Simmias and others, and he will have no difficulty in finding friends in Thessaly and other places. Socrates is afraid that Crito is but pressing upon him the opinions of the many:  whereas, all his life long he has followed the dictates of reason only and the opinion of the one wise or skilled man. There was a time when Crito himself had allowed the propriety of this. And although some one will say 'the many can kill us,' that makes no difference; but a good life, in other words, a just and honourable life, is alone to be valued. All considerations of loss of reputation or injury to his children should be dismissed:  the only question is whether he would be right in attempting to escape. Crito, who is a disinterested person not having the fear of death before his eyes, shall answer this for him. Before he was condemned they had often held discussions, in which they agreed that no man should either do evil, or return evil for evil, or betray the right. Are these principles to be altered because the circumstances of Socrates are altered? Crito admits that they remain the same. Then is his escape consistent with the maintenance of them? To this Crito is unable or unwilling to reply. Socrates proceeds:--Suppose the Laws of Athens to come and remonstrate with him:  they will ask 'Why does he seek to overturn them?' and if he replies, 'they have injured him,' will not the Laws answer, 'Yes, but was that the agreement? Has he any objection to make to them which would justify him in overturning them? Was he not brought into the world and educated by their help, and are they not his parents? He might have left Athens and gone where he pleased, but he has lived there for seventy years more constantly than any other citizen.' Thus he has clearly shown that he acknowledged the agreement, which he cannot now break without dishonour to himself and danger to his friends. Even in the course of the trial he might have proposed exile as the penalty, but then he declared that he preferred death to exile. And whither will he direct his footsteps? In any well-ordered state the Laws will consider him as an enemy. Possibly in a land of misrule like Thessaly he may be welcomed at first, and the unseemly narrative of his escape will be regarded by the inhabitants as an amusing tale. But if he offends them he will have to learn another sort of lesson. Will he continue to give lectures in virtue? That would hardly be decent.   [...]

\noindent\textbf{Question}
Who persuades Socrates to escape?

\noindent\textbf{Ground Truth}
Crito, Crito convinces Socrates to leave his cell.

\noindent\textbf{\MODEL~Output}
Crito {\color{green}\ding{51}}

\noindent\textbf{\llama~Output}
According to the text, it is Crito who persuades Socrates to escape. {\color{green}\ding{51}}

\noindent\textbf{ChatGPT-3.5 Output}
Crito {\color{green}\ding{51}}

\end{example}

In this example, while all models provide correct answers, \llama~responds in a more verbose manner, deviating from the given instruction.

\newpage

\begin{example}{Qasper}
\noindent\textbf{Instruction}
You are given a scientific article and a question. Answer the question as concisely as you can, using a single phrase or sentence if possible. If the question cannot be answered based on the information in the article, write ``unanswerable''. If the question is a yes/no question, answer ``yes'', ``no'', or ``unanswerable''. Do not provide any explanation.

\noindent\textbf{Document}
[...] For all the features, we separated the words in the tweets based on the language annotation (Section 2.3) and prepared the feature vector for each tweet by combining the vectors for both the languages . Previous researches shows that letter n-grams are very efficient for classifying text. They are language independent and does not require expensive text pre-processing techniques like tokenization, stemming and stop words removal, hence in the case of code-mix texts, this could yield good results BIBREF16 , BIBREF17 . Since the number of n-grams can be very large we took trigrams which occur more than ten times in the corpus. For classifying humor in texts, it is important to understand the semantics of the sentence. Thus, we took a three word window as a feature to train our classification models to incorporate the contextual information. Many jokes and idioms sometimes have common words. We identified those words and took them as as a feature for classification. In the preprocessing step, we decomposed hashtags using camel cases and added them along with the words. Hence, common words in the hashtags were also included in the feature vector. \hl{Classification Approach and Results\par We experimented with four different classifiers, namely, support vector machine BIBREF18 , random forest, extra tree and naive bayes classifier BIBREF19 .} Chi square feature selection algorithm is applied to reduces the size of our feature vector. For training our system classifier, we used Scikit-learn BIBREF19 . 10-fold cross validation on 3543 code-mixed tweets was carried out by dividing the corpus into 10 equal parts with nine parts as training corpus and rest one for testing. Mean accuracy is calculated by taking the average of the accuracy obtained in each iteration of the testing process. Table 2 shows the accuracy for each feature when trained using mentioned classifiers along with the accuracy when all the features are used along with the overall accuracy. Support vector machine with radial basis function kernel and extra tree classifier performs better than other classifiers and yields 69.3\% and 67.8\% accuracy respectively. The reason kernel SVM yields the best result is because the number of observations is greator than the number of features BIBREF20 . N-grams proved to be the most efficient in all classification models followed by common words and hastags. Bag-of-words feature performed the worst in SVM, random forest and extra tree classifier but yielded better result in naive bayes classifiers. Accuracies mentioned in table 2 were calculated using fine tuning of model parameters using grid search. Conclusion and Future Work\par In this paper, we describe a freely available corpus of 3453 English-Hindi code-mixed tweets. The tweets are annotated with humorous(H) and non-humorous(N) tags along with the language tags at the word level. The task of humor identification in social media texts is analyzed as a classification problem and several machine learning classification models are used. \hl{The features used in our classification system are n-grams, bag-of-words, common words and hashtags.} N-grams when trained with support vector machines with radial basis function kernel performed better than other features and yielded an accuracy of 68.5\%. The best accuracy (69.3\%) was given by support vector machines with radial basis function kernel. This paper describes the initial efforts in automatic humor detection in code-mixed social media texts. Corpus can be annotated with part-of-speech tags at the word level which may yield better results in language detection.  [...]

\noindent\textbf{Question}
What type of system does the baseline classification use?

\noindent\textbf{Ground Truth}
support vector machine BIBREF18 , random forest, extra tree and naive bayes classifier BIBREF19, Classification system use n-grams, bag-of-words, common words and hashtags as features and SVM, random forest, extra tree and NB classifiers.

\noindent\textbf{\MODEL~Output}
The baseline classification uses n-grams, bag-of-words, common words and hashtags. {\color{green}\ding{51}}

\noindent\textbf{\llama~Output}
to the article, the baseline classification system uses a support vector machine (SVM) classifier. {\color{red}\ding{55}}

\noindent\textbf{ChatGPT-3.5 Output}
Supervised classification system {\color{red}\ding{55}}
\end{example}

\newpage

\begin{example}{Musique}
\noindent\textbf{Instruction}
You are given several paragraphs from Wikipedia and a question. Answer the question as concisely as you can, using a single phrase if possible. If the question cannot be answered based on the information in the paragraphs, write ``unanswerable''. Do not provide any explanation.

\noindent\textbf{Document}
[...] Title: Israel\par Israel, officially the State of Israel, is a country in the Middle East, on the southeastern shore of the Mediterranean Sea and the northern shore of the Red Sea. It has land borders with Lebanon to the north, Syria to the northeast, Jordan on the east, the Palestinian territories of the West Bank and Gaza Strip to the east and west, respectively, and Egypt to the southwest. The country contains geographically diverse features within its relatively small area. Israel's economy and technology center is Tel Aviv, while its seat of government and proclaimed capital is Jerusalem, although the state's sovereignty over East Jerusalem is not recognised internationally. The population of Israel was estimated in 2017 to be 8,777,580 people, of whom 74.7\% were Jewish, 20.8\% Arab and 4.5\% others.\par Title: The World (archipelago)\par The World or The World Islands, (``Juzur al-Ālam'') is an artificial archipelago of various small islands constructed in the rough shape of a world map, located in the waters of the Persian Gulf, off the coast of Dubai, United Arab Emirates. The World islands are composed mainly of sand dredged from Dubai's shallow coastal waters, and are one of several artificial island developments in Dubai. The World's developer is Nakheel Properties, and the project was originally conceived by Sheikh Mohammed bin Rashid Al Maktoum, the ruler of Dubai. The actual construction was done by two Dutch (joint venture) specialist companies, Van Oord and Boskalis. The same companies also created the Palm Jumeirah.\par Title: Israel\par Israel is considered the most advanced country in Southwest Asia and the Middle East in economic and industrial development. Israel's quality university education and the establishment of a highly motivated and educated populace is largely responsible for spurring the country's high technology boom and rapid economic development. In 2010, it joined the OECD. The country is ranked 3rd in the region and 38th worldwide on the World Bank's Ease of Doing Business Index as well as in the World Economic Forum's Global Competitiveness Report. It has the second-largest number of startup companies in the world (after the United States) and the largest number of NASDAQ-listed companies outside North America.\par Title: Israel\par Israel is one of the world's technological leaders in water technology. In 2011, its water technology industry was worth around \$2 billion a year with annual exports of products and services in the tens of millions of dollars. The ongoing shortage of water in the country has spurred innovation in water conservation techniques, and a substantial agricultural modernization, drip irrigation, was invented in Israel. [...]

\noindent\textbf{Question}
When was the region that lies to the north of the region Israel is located and where The World is located created?

\noindent\textbf{Ground Truth}
1930

\noindent\textbf{\MODEL~Output}
1946 {\color{red}\ding{55}}

\noindent\textbf{\llama~Output}
The region that lies to the north of Israel was created in the mid-18th century. {\color{red}\ding{55}}

\noindent\textbf{ChatGPT-3.5 Output}
unanswerable {\color{red}\ding{55}}
\end{example}

This example underscores that the pertinent information for the given question cannot be discerned within the context, indicating that certain examples may not have been accurately constructed.
The paragraphs do not provide information to answer when the region north of Israel and The World archipelago was created. 
There is information about Israel and The World islands located in Dubai, but no mention of the region to the north of Israel.
The \MODEL~and \llama~models manifest hallucinations, whereas GPT answers accurately following the given instruction.

\newpage

\begin{example}{Musique}
\noindent\textbf{Instruction}
You are given several paragraphs from Wikipedia and a question. Answer the question as concisely as you can, using a single phrase if possible. If the question cannot be answered based on the information in the paragraphs, write ``unanswerable''. Do not provide any explanation.

\noindent\textbf{Document}
Title: Jerome Quinn\par Born in Green Bay, Wisconsin, Quinn was a realtor and served on the Green Bay Common Council, the Brown County, Wisconsin Board of Supervisors, the local Board of Education, and the Wisconsin State Assembly from 1955 until 1973. He was a Republican.\par Title: Max Rosenthal Tenement in Bydgoszcz\par Max Rosenthal House is a historical tenement located at Gdańska Street N°42 in downtown Bydgoszcz, Poland, built when the city was part of the Kingdom of Prussia. It is registered on the Kuyavian-Pomeranian Voivodeship Heritage List.\par Title: Borders of China\par China shares international borders with 14 sovereign states. In addition, there is a 30 - km border with the special administrative region of Hong Kong, which was a British dependency before 1997, and a 3 km border with Macau, a Portuguese territory until 1999. With a land border of 22,117 kilometres (13,743 mi) in total it also has the longest land border of any country.\par Title: Notogawa, Shiga\par Notogawa Station (Location: N35.179899,E136.165913) is the only Japan Railway station in Higashiomi. The station is a rapid stop on the JR Biwako Line, located between stations in Omi-Hachiman to the east and Hikone to the west. The town shares a small border with Lake Biwa to the northwest.\par Title: Tenement at Gdanska street 75, Bydgoszcz\par Tenement at Gdanska street 75 is a historical habitation building located at Gdańska Street N°75, in Bydgoszcz. It is registered on the Kuyavian-Pomeranian Voivodeship Heritage List.\par Title: Tenement (band)\par \hl{Tenement is a three piece rock band from Appleton, Wisconsin formed in 2006.} They are often associated with the American hardcore punk scene. Their recorded output has been described as everything from ``noise pop'' to ``fuzz punk'', while in a live setting they are often known for experimentation, improvisation, and high-energy performance. The visual art of singer/guitarist Amos Pitsch is associated with most of their records, as well as several records by other notable punk and hardcore bands. In 2016, they were included in ``Rolling Stone's'' ``10 Great Modern Punk Bands''.\par Title: Pulaski High School\par Pulaski High School is a public high school in Pulaski, Wisconsin, in Brown County, Wisconsin (school district also serves parts of Shawano, Outagamie and Oconto counties), that serves students in grades 9 through 12. Its mascot is the Red Raider.[...]

\noindent\textbf{Question}
What is the capital of the county that shares a border with the county where the band Tenement was formed?

\noindent\textbf{Ground Truth}
Green Bay

\noindent\textbf{\MODEL~Output}
Appleton {\color{red}\ding{55}}

\noindent\textbf{\llama~Output}
Appleton. {\color{red}\ding{55}}

\noindent\textbf{ChatGPT-3.5 Output}
Appleton, Wisconsin {\color{red}\ding{55}}
\end{example}

This example shows that the ground truth is not correct.
The paragraph states that the band Tenement is ``a three piece rock band from Appleton, Wisconsin formed in 200''. 
There is no information about the capital of the county that shares a border with Appleton in the document but Appleton is the closest answer given the context.

\end{document}